\documentclass[pdflatex,iicol,sn-mathphys-ay]{sn-jnl}

\usepackage{graphicx}
\usepackage{amsmath,amssymb,amsfonts}
\usepackage{array}
\usepackage{tabularx}
\usepackage{textcomp}
\usepackage{url}
\usepackage{verbatim}
\usepackage{indentfirst}
\usepackage{multirow}
\usepackage{pifont}
\usepackage{booktabs}
\usepackage{colortbl}
\let\orcidlogo\relax
\usepackage{orcidlink}
\usepackage{adjustbox}
\usepackage{arydshln}

\definecolor{imposerow}{gray}{0.92}
\newcommand{\imposetablestyle}{%
  \footnotesize
  \renewcommand{\arraystretch}{1.18}%
  \setlength{\tabcolsep}{4pt}}

\providecommand{\Checkmark}{\ding{51}}
\providecommand{\XSolidBrush}{\ding{55}}

\makeatletter
\newcommand{\tablegroupstrut}{\rule{0pt}{\dimexpr\ht\@arstrutbox+2pt\relax}}
\let\table\tableorg
\let\endtable\endtableorg

\let\ijcvoriginaltabular\tabular
\let\ijcvoriginalendtabular\endtabular
\newcommand{\ijcvtabular}[1]{%
  \begin{adjustbox}{max width=\columnwidth}%
  \ijcvoriginaltabular{#1}}
\newcommand{\ijcvendtabular}{%
  \ijcvoriginalendtabular
  \end{adjustbox}}
\AtBeginEnvironment{table}{%
  \let\tabular\ijcvtabular
  \let\endtabular\ijcvendtabular}
\AtBeginEnvironment{tabularx}{%
  \let\tabular\ijcvoriginaltabular
  \let\endtabular\ijcvoriginalendtabular}
\makeatother

\let\snabstract\abstract
\renewcommand{\abstract}{}

\begin{document}

\title[IMPose: Closed-Loop Correction Propagation for Interactive Multi-Person Pose Annotation]
{IMPose: Closed-Loop Correction Propagation for Interactive Multi-Person Pose Annotation}

\author[1]{\fnm{Haoyang} \sur{Ge}\orcidlink{0009-0005-2839-3744}}
\equalcont{Equal contribution.}

\author[1]{\fnm{Jian} \sur{Ma}\orcidlink{0000-0002-9269-1130}}
\equalcont{Equal contribution.}

\author[1]{\fnm{Ziwen} \sur{Wang}\orcidlink{0009-0006-6348-5182}}

\author[1]{\fnm{Qihe} \sur{Wang}\orcidlink{0009-0008-1055-1478}}

\author[1]{\fnm{Jianqi} \sur{Fan}\orcidlink{0009-0001-1997-4820}}

\author[3]{\fnm{Hongzhi} \sur{Yu}\orcidlink{}}

\author[2]{\fnm{Xingyu} \sur{Chen}\orcidlink{0000-0003-3627-0371}}

\author*[1]{\fnm{Kun} \sur{Li}\orcidlink{0000-0003-2326-0166}}
\email{lik@tju.edu.cn}

\affil[1]{\orgname{Tianjin University}, \orgaddress{\city{Tianjin}, \postcode{300350}, \country{China}}}
\affil[2]{\orgname{Zhongguancun Academy}, \orgaddress{\city{Beijing}, \postcode{100094}, \country{China}}}
\affil[3]{\orgname{Tiandy}, \orgaddress{\city{Tianjin}, \country{China}}}

\snabstract{\begin{abstract}

High-quality dynamic human pose annotations provide important supervision for learning fine-grained human motion, yet dense multi-person video annotation remains labor-intensive. Sparse human corrections can reduce this effort, but their propagated locations may become unreliable and ambiguous across identities. Existing interactive refinement and temporal propagation methods do not explicitly address how imperfect propagated corrections should participate in subsequent multi-person pose inference. We propose IMPose, a correction-conditioned temporal inference framework for interactive multi-person pose annotation. IMPose separates temporal correspondence from pose estimation: propagated corrections are treated as geometric priors rather than final predictions and are refined by a correction-aware decoder using target-frame visual evidence. A keypoint-aware identity state further combines structured pose geometry with trajectory history to preserve person- and joint-level correspondence. The refined pose and identity state are fed back to subsequent frames, forming a closed temporal correction loop. Under a controlled sparse-correction protocol on PoseTrack21, one correction improves mean average precision (mAP) from 38.87 to 67.99 and Higher Order Tracking Accuracy (HOTA) from 52.21 to 64.89. At a 20-frame correction interval, IMPose achieves 89.03\% keypoint-level Object Keypoint Similarity (K-OKS), compared with 81.41\% for CoTracker and 70.12\% for PoseWarper-based propagation. We further extend PoseTrack21 with 188K additional person instances through 3.55M annotation updates to keypoints, bounding boxes, and person IDs using 100 annotator-hours. The annotation tool, code, and extended dataset will be released.


\end{abstract}

}


\maketitle

\begin{figure*}[!t]
    \centering
    \includegraphics[width=\textwidth]{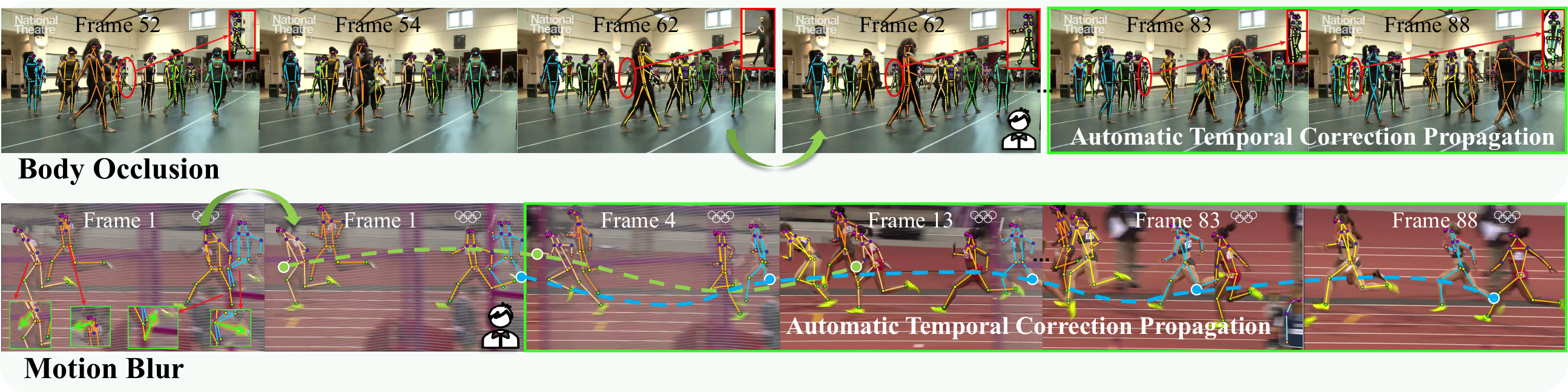}
    \caption{\textbf{Demonstrations of IMPose Annotating Videos Containing Multiple Persons}. The first row demonstrates that complete disappearance of a subject in frame 54 leads to detection failure upon reappearance in frame 62. The second row highlights pose estimation errors in multi-athlete scenes caused by motion blur. IMPose effectively corrects these anomalies and propagates temporal corrections, achieving reliable keypoint propagation across up to 88 subsequent frames in challenging scenarios. Zoom in for a better view.}
    \label{fig:fig1}
\end{figure*}

\section{Introduction}

High-quality human pose annotations provide important supervision for video understanding tasks such as action recognition, motion analysis, and VR/AR interaction. In multi-person videos, however, dense pose annotation remains labor-intensive because annotators must jointly verify keypoint locations, missed persons, bounding boxes, and identity assignments across time. The challenge is therefore not only to estimate accurate poses in individual frames, but also to efficiently incorporate sparse human corrections into temporally coherent and identity-consistent pose trajectories.

Existing pose annotation systems largely follow an open-loop paradigm. They first generate pose predictions using image-level pose estimators~\cite{yang2023effective} and then associate person instances across frames using tracking modules~\cite{wojke2017simple,poseflow}, as in X-AnyLabeling~\cite{wang2023advanced}. Recent video pose estimation and tracking methods further improve temporal modeling and person association~\cite{he2024decoupled,feng2025high,chen2025causal,alphapose,doering2023gated}. These advances provide increasingly accurate initial trajectories, but human corrections are typically applied after model inference. Once an annotator corrects a shifted keypoint, a missed person, or an identity error, the corrected information does not necessarily participate in subsequent pose inference. Similar errors may therefore recur in neighboring frames and require repeated intervention.

Human-in-the-loop methods reduce this burden by exploiting sparse user input in different ways. Click-Pose~\cite{yang2023neural} incorporates user clicks into pose refinement within individual images, while PoseWarper~\cite{posewarper} transfers sparse pose evidence across frames through pose-aware temporal warping. More recently, iPose~\cite{liu2024ipose} propagates user-edited 2D joint handles using CoTracker and incorporates the resulting constraints into the optimization of subsequent 3D poses under temporal consistency. However, Click-Pose supports interactive correction in static images without propagating edits across frames. PoseWarper propagates pose evidence across frames but does not provide an interactive workflow for correcting person IDs. iPose focuses on editing one person's 3D motion rather than jointly maintaining pose annotations and identities for multiple people. Our focus is interactive, identity-consistent pose annotation in dynamic multi-person videos. This task presents two challenges: using imperfect propagated corrections to improve target-frame pose estimation and keeping the resulting poses associated with the correct people over time.

A human correction provides reliable evidence about a particular person's joint in the edited frame, but its propagated location can become inaccurate under motion, occlusion, blur, deformation, and person interaction. Using the propagated location directly can carry correspondence errors into later annotations, while estimating subsequent poses from images alone leaves the earlier correction unused. The first challenge is to preserve the information supplied by human correction while allowing the target-frame image to revise its propagated location. We therefore formulate interactive multi-person pose annotation as a correction-conditioned temporal inference problem. We separate temporal correspondence from pose estimation so that propagated corrections guide where to search without prescribing the target-frame pose. The propagated correction enters subsequent pose inference as a geometric prior. The pose decoder then reconciles this prior with current-frame visual evidence to determine the target-frame pose. In this way, temporal propagation provides candidate correspondence, while pose estimation remains grounded in the observation of each new frame.

Based on this formulation, we propose IMPose, a correction-conditioned temporal inference framework for interactive multi-person pose annotation. Its pose decoder is trained with imperfect keypoint references and simulated sparse human corrections, enabling it to refine partially reliable geometric priors using image evidence. At inference time, human corrections propagated from previous frames instantiate such priors in subsequent frames. Instead of being directly accepted as annotations, they are converted into box and keypoint reference points that condition the pose decoder and are refined according to the target-frame observation. This reference-point formulation connects correction-aware learning with temporal propagation, allowing the decoder to refine propagated human feedback using the same geometric-prior representation employed during training.

Second, accurate pose refinement alone does not resolve person identity: when people overlap or exchange positions, a plausible pose can still be associated with the wrong trajectory. We therefore propose a keypoint-aware identity state that combines structured pose geometry with identity-conditioned trajectory history. A keypoint-aware pose embedding captures joint configuration and relative body geometry, while a trajectory bank maintains the temporal state of each person. After a propagated correction is refined, the updated pose and identity state are written back to the trajectory bank and used in subsequent frames. The trajectory bank therefore carries refined pose representations into subsequent identity association, linking pose refinement and person tracking within the temporal correction loop.

Experiments on PoseTrack21~\cite{doering2022posetrack21} and 3DPW demonstrate the effectiveness of IMPose under sparse interaction, and a manual annotation study shows reduced annotation time. We further use IMPose to expand PoseTrack21 with 187,920 additional person instances through 3,548,968 annotation updates to keypoints, bounding boxes, and person IDs using 100 annotator-hours. Downstream experiments show that training DSTA and fine-tuning CameraHMR with IMPose-corrected annotations improve 2D keypoint localization compared with using the original pseudo-labels.

Our contributions are summarized as follows:
\begin{itemize}
    \item We propose IMPose, a framework for interactive, identity-consistent pose annotation in dynamic multi-person videos, in which we formulate annotation as correction-conditioned temporal inference.

    \item We propose a correction-conditioned refinement mechanism that connects temporal propagation with pose decoding through box and keypoint reference priors, allowing target-frame visual evidence to revise imperfect propagated locations.

    \item We introduce a keypoint-aware identity state that combines structured pose geometry with trajectory history, allowing refined pose representations to inform subsequent person association.

    \item We develop an interactive annotation platform for correcting keypoints, bounding boxes, and person identities, with corrections propagated to subsequent frames. The platform, code, and extended annotations will be publicly released.
\end{itemize}

\section{Related Work}
\label{sec:related_work}

\subsection{Open-loop Pose Annotation}

Open-loop pose annotation systems reduce manual effort by applying pose estimators and tracking algorithms to unlabeled videos. General annotation platforms such as AnyLabeling~\cite{anylabeling}, X-AnyLabeling~\cite{wang2023advanced}, and CVAT~\cite{cvat} integrate modern vision models, including SAM2~\cite{ravi2024sam}, to generate initial labels and support video annotation workflows. In our evaluation, X-AnyLabeling serves as an automatic annotation baseline with DWPose-based pose initialization~\cite{yang2023effective}. Such pipelines typically combine image-level pose estimation with tracking algorithms such as DeepSORT~\cite{wojke2017simple} or PoseFlow~\cite{poseflow}. The resulting pipeline first predicts poses in individual frames and then associates person instances across time. FairMOT~\cite{zhang2021fairmot} jointly learns object detection and re-identification features for multi-object tracking.

This paradigm has benefited from rapid progress in multi-person pose estimation and pose tracking. Top-down methods, including AlphaPose~\cite{alphapose}, HRNet~\cite{hrnet}, and ViTPose~\cite{vitpose}, estimate poses after detecting each person, while bottom-up methods such as OpenPose~\cite{openpose}, Associative Embedding~\cite{associative_embedding}, and PifPaf~\cite{pifpaf} detect joints and group them into person instances. \citet{zhang2021towards} exploit spatial and channel context to improve human keypoint detection, including the localization of invisible joints. Video-based methods further improve temporal consistency through frame-to-frame association or temporal pose modeling, including PoseTrack~\cite{posetrack}, Detect-and-Track~\cite{detect_and_track}, VideoPose3D~\cite{videopose3d}, DSTA~\cite{he2024decoupled}, GLSMamba~\cite{feng2025high}, CM-Pose~\cite{chen2025causal}, and Gated Attention Transformer~\cite{doering2023gated}. \citet{ranjan2020learning} learn human-specific optical flow to capture articulated motion and occlusion in multi-person scenes. These methods provide stronger pose trajectories and are valuable components of automatic annotation systems.

When used for annotation, however, these approaches typically operate in an open-loop manner. The model first produces pose and identity predictions, after which annotators inspect and correct the outputs. Human corrections are not explicitly incorporated into subsequent pose inference, and therefore do not become part of the temporal inference state. As a result, errors caused by occlusion, motion blur, missed detections, or person interactions may recur in nearby frames and require repeated manual correction.

\subsection{Human-in-the-loop Pose Annotation}

Human-in-the-loop annotation reduces labeling cost by incorporating sparse human input into automatic prediction and refinement. One line of work relies on sparse manual annotation followed by temporal propagation. VATIC~\cite{vondrick2013efficiently} combines manual keyframe annotation with automatic interpolation to reduce the cost of video labeling. EPIC-KITCHENS VISOR~\cite{darkhalil2022epic} reduces video annotation effort by labeling selected frames and inferring intermediate ones. PoseWarper~\cite{posewarper} follows a similar principle for human pose estimation by learning pose-aware temporal warping from sparsely labeled videos and transferring pose evidence from labeled frames to nearby unlabeled frames. More recently, STDPose~\cite{jiao2025spatiotemporal} learns spatiotemporal representations from sparsely labeled videos for pose propagation and estimation. These approaches exploit temporal structure to extend sparse annotations across frames.

Another line of work focuses on interactive pose refinement. Click-Pose~\cite{yang2023neural} incorporates sparse user corrections into pose prediction and learns to refine inaccurate or partially corrected poses within individual images. PoseFix~\cite{posefix} similarly demonstrates that common pose estimation errors can be corrected through learned refinement. These methods establish the effectiveness of correction-aware pose refinement, but mainly operate within individual frames and do not address how corrected pose information should participate in inference across time.

Recent methods extend interactive correction to video. iPose~\cite{liu2024ipose} propagates user-edited 2D joint handles across frames using CoTracker and incorporates the propagated constraints into the optimization of subsequent 3D poses and camera parameters. Its interaction focuses on editing one person's 3D motion rather than jointly maintaining pose annotations and identities for multiple people. General point trackers such as CoTracker~\cite{karaev24cotracker3} further provide accurate temporal correspondences for propagating user-specified points. For animal pose annotation, PosePAL~\cite{pan2026posepal} uses sparse annotations to adapt a point tracker through test-time optimization, generating dense labels across video frames. These studies show that sparse human corrections can be transferred across time and subsequently exploited for pose refinement.

\begin{figure*}[t!]
  \centering
  \includegraphics[width=\textwidth]{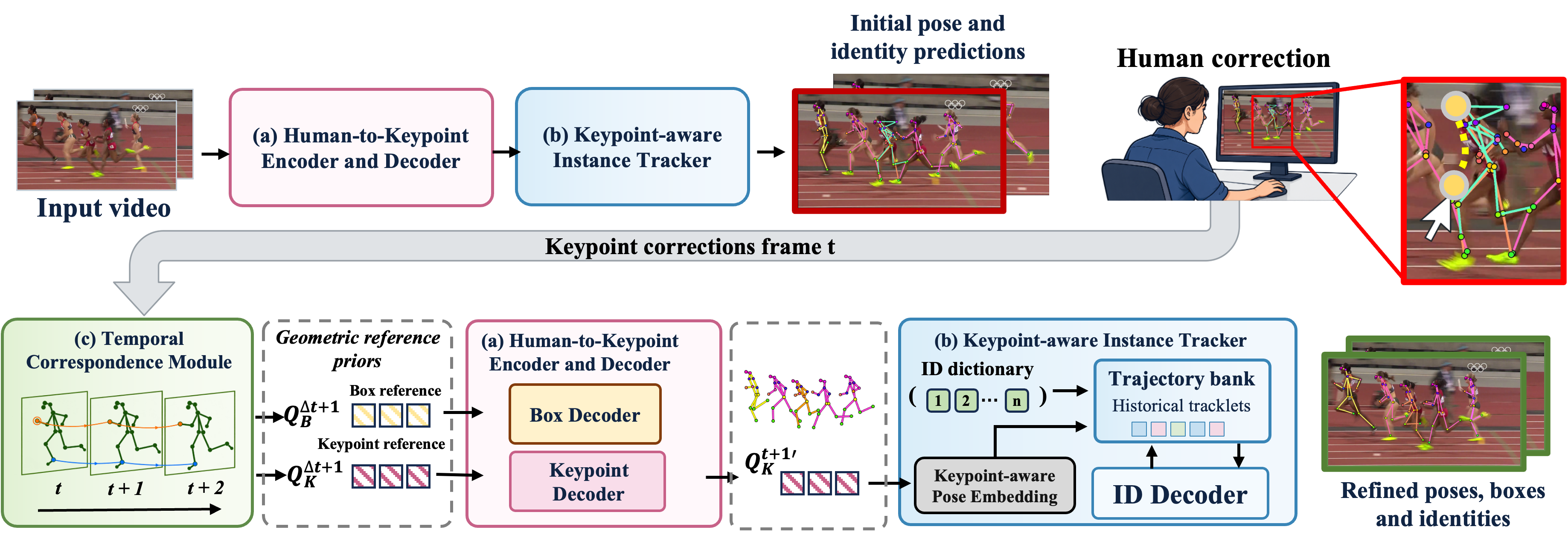}
  \caption{\textbf{Overview of IMPose.}
  The Human-to-Keypoint encoder and decoder (a) and keypoint-aware instance tracker (b) generate initial poses and identities for annotator review.
  After a keypoint is corrected at frame $t$, the temporal correspondence module (c) propagates its location to subsequent frames.
  These propagated locations serve as geometric priors for box and keypoint decoding in (a), where target-frame visual evidence refines the pose.
  Keypoint-aware pose embeddings and the trajectory bank in (b) support identity association.
  Refined pose representations and identity assignments update the trajectory bank for subsequent frames, forming a closed temporal correction loop.}
  \label{fig: Fig:3.1}
\end{figure*}

We propose IMPose for interactive, identity-consistent 2D pose annotation in dynamic multi-person videos. We treat propagated corrections as box and keypoint reference priors and refine them through a correction-aware pose decoder using target-frame visual evidence. Our keypoint-aware identity state combines structured pose geometry with trajectory history. Refined pose representations and identity assignments are written back to the trajectory bank for subsequent person association, forming a closed temporal correction loop.

\section{Method}

\subsection{Overview}
\label{sec:Overviewr}

We present IMPose, a correction-conditioned framework for interactive multi-person pose annotation, as illustrated in Fig.~\ref{fig: Fig:3.1}. 
The key idea is to separate \emph{temporal correspondence} from \emph{pose inference}. 
A human correction provides reliable local evidence in the edited frame, whereas its location after temporal propagation remains uncertain due to motion, occlusion, blur, and interactions with nearby people. 
Instead of directly accepting a propagated coordinate as the target-frame annotation, IMPose treats it as an imperfect geometric prior and re-estimates the pose from the current-frame visual evidence. 
The refined pose is further associated with a keypoint-aware identity state, allowing the correction to retain its person- and joint-level semantics over time.

Given a video with $T$ frames
$\mathbf{I}=\{\mathbf{I}^{1},\mathbf{I}^{2},\ldots,\mathbf{I}^{T}\}$,
a Human-to-Keypoint encoder and decoder first estimate multi-person bounding boxes
$\mathbf{B}=\{\mathbf{B}^{1},\mathbf{B}^{2},\ldots,\mathbf{B}^{T}\}$
and keypoints
$\mathbf{K}=\{\mathbf{K}^{1},\mathbf{K}^{2},\ldots,\mathbf{K}^{T}\}$.
A keypoint-aware instance tracker then assigns person identities by combining instance features, structured pose geometry, and historical trajectory information.

When an annotator corrects the prediction at frame $t$, the corrected boxes and keypoints are propagated to subsequent frames. 
Rather than being used as final predictions, the propagated locations are converted into box and keypoint reference priors,
$\mathbf{Q}_{B}^{\Delta t}$ and $\mathbf{Q}_{K}^{\Delta t}$.
For a target frame $t+\Delta t$, these priors condition the box and keypoint decoders, which refine the candidate locations according to the visual evidence of the target frame.
The resulting pose is subsequently associated with the keypoint-aware identity state and written back to the trajectory bank, providing the state for later frames.

The resulting inference process can be summarized as
\begin{equation}
\mathbf{C}_{t}
\xrightarrow{\mathcal{T}}
\widetilde{\mathbf{C}}_{t\rightarrow t+\Delta t}
\xrightarrow{\mathcal{D}(\mathbf{I}^{t+\Delta t})}
\hat{\mathbf{P}}_{t+\Delta t}
\xrightarrow{\mathcal{A}}
\mathbf{S}_{t+\Delta t},
\end{equation}
where $\mathbf{C}_{t}$ denotes a human correction,
$\mathcal{T}$ denotes temporal correspondence,
$\widetilde{\mathbf{C}}_{t\rightarrow t+\Delta t}$ is the propagated correction prior,
$\mathcal{D}$ denotes image-conditioned pose refinement,
$\hat{\mathbf{P}}_{t+\Delta t}$ is the refined target-frame pose,
and $\mathbf{S}_{t+\Delta t}$ is the updated pose--identity state.
The updated state is further used in subsequent frames, forming a closed temporal correction loop.

\subsection{Keypoint-Aware Identity State}
\label{sec:Instance-level Tracking}

A propagated correction is useful for multi-person annotation only when it remains associated with the same person and semantic joint.
Coordinate correspondence alone is insufficient for this purpose, particularly when people overlap, disappear, or exchange positions.
IMPose therefore maintains a keypoint-aware identity state that combines instance-level features with structured pose geometry and historical trajectory information.
This state provides the person- and joint-level semantics required to interpret propagated corrections consistently across frames.

\textbf{Multi-person pose representation.}
For each frame $\mathbf{I}^{t}$, we adopt a DETR-style Human-to-Keypoint backbone to obtain person and pose representations, as shown in Fig.~\ref{fig: Fig:3.1}(a).
A Deformable Transformer Encoder~\cite{zhu2020deformable} extracts multi-scale visual features $\mathbf{F}^{t}$.
Following the query selection strategy of DINO~\cite{zhang2022dino}, human content queries and box queries,
denoted by $\mathbf{Q}_{C}^{t}$ and $\mathbf{Q}_{B}^{t}$,
are selected from the encoded features.
The Box Decoder first updates the box queries to obtain candidate human regions $\mathbf{Q}_{B}^{t'}$.
These queries are then combined with learnable keypoint embeddings $\mathbf{Q}_{K}^{t}$ and passed to the Keypoint Decoder, producing refined box queries $\mathbf{Q}_{B}^{t''}$ and keypoint queries $\mathbf{Q}_{K}^{t'}$.
The corresponding predictions are mapped to bounding boxes $\mathbf{B}^{t}$ and keypoints $\mathbf{K}^{t}$.

\textbf{Keypoint-aware pose embedding.}
To complement appearance-based instance representation with pose semantics, we construct a structured embedding from the predicted keypoints.
For each detected person, the $17$ keypoints are flattened into a coordinate representation and augmented with relative geometric features.
For every valid keypoint pair $(i,j)$ defined by the human skeleton adjacency matrix, we compute
the relative vector
$\mathbf{v}_{ij}=\mathbf{p}_{i}-\mathbf{p}_{j}$,
its Euclidean distance $\mathbf{R}^{t}_{\mathrm{dist}}$,
and normalized direction $\mathbf{R}^{t}_{\mathrm{dir}}$.
We further map the joint coordinates into Fourier features
$\mathbf{F}_{\mathbf{K}}^{t}$
to improve spatial discrimination.
The resulting keypoint-aware embedding is
\begin{equation}
\mathbf{E}_{\mathbf{K}^{t}}
=
f_{\mathrm{p}}
\left(
\operatorname{concat}
\left(
\mathbf{K}^{t},
\mathbf{R}^{t},
\mathbf{F}_{\mathbf{K}}^{t}
\right)
\right),
\end{equation}
where $f_{\mathrm{p}}$ denotes a linear projection.

Unlike a generic point correspondence, $\mathbf{E}_{\mathbf{K}^{t}}$ represents the structured configuration of a human pose.
It therefore provides semantic information for distinguishing, for example, the left knee of one person from a nearby image point belonging to another person.

\textbf{Trajectory bank and identity association.}
To maintain temporal identity consistency, IMPose stores box features, keypoint-aware pose embeddings, and identity prompts in a trajectory bank, as shown in Fig.~\ref{fig: Fig:3.1}(b).
Following the ID dictionary formulation~\cite{gao2025multiple}, we define a set of learnable identity embeddings
$\mathbf{Z}=\{\mathbf{z}^{1},\mathbf{z}^{2},\ldots,\mathbf{z}^{n}\}$.
For a detected person in frame $t$, its tracklet representation is
\begin{equation}
\mathbf{T}^{t}
=
\operatorname{concat}
\left(
\mathbf{Q}_{B}^{t''},
\mathbf{E}_{\mathbf{K}^{t}},
\mathbf{z}^{n}
\right).
\end{equation}

The historical trajectory state is denoted by
\begin{equation}
\mathbf{T}^{t-T:t-1}
=
\{
\mathbf{T}^{t-T},\ldots,\mathbf{T}^{t-1}
\}.
\end{equation}
An ID Decoder, implemented as a transformer decoder, attends from current tracklets to the historical trajectory state and predicts identity assignments.
When multiple tracklets are assigned to the same identity, the highest-scoring match is retained, while unmatched tracklets initialize new trajectories.

Importantly, the trajectory bank is also part of the correction loop.
After a propagated correction is refined, its updated pose representation and identity assignment are written back to the trajectory bank.
Subsequent frames therefore inherit not only a geometric correction prior but also the person-level and joint-level state required to interpret that prior consistently.
In this way, correction propagation and identity association are coupled through a shared evolving temporal state.

\subsection{Correction-Conditioned Temporal Pose Refinement}
\label{sec:keypoint-level Temporal Correction Propagation Mechanism}

The central problem is not merely how to propagate a corrected keypoint across frames, but how to use an imperfect propagated correction to improve pose estimation in a subsequent frame.
A correction at frame $t$ provides reliable local evidence, whereas its propagated position at frame $t+\Delta t$ is only an estimate of where the joint may appear.
IMPose therefore decouples temporal correspondence from pose estimation:
the temporal correspondence module proposes candidate joint locations, while the pose decoder treats these locations as geometric priors and refines them according to the target-frame visual evidence.

\textbf{Correction-aware pose refinement learning.}
A critical property of the pose decoder is that it is explicitly trained to operate on imperfect and partially corrected keypoint priors.
Following the pose error modeling strategy of Click-Pose~\cite{yang2023neural}, we randomly perturb ground-truth keypoint references during training to construct noisy pose priors.
A subset of the keypoints remains unperturbed, so that the decoder observes a mixture of reliable and inaccurate geometric references.
We further simulate sparse human interaction by identifying erroneous predicted keypoints, replacing selected joints with their ground-truth locations, and supervising the decoder to refine the complete pose.

This training does not assume that an input reference is already the correct pose.
Instead, it learns a correction-aware refinement function that reconciles partially reliable geometric priors with image evidence.
Conceptually, the training process can be written as
\begin{equation}
\widetilde{\mathbf{K}}^{t}
=
\mathcal{E}
\left(
\mathbf{K}^{t};
\boldsymbol{\xi}
\right),
\end{equation}
where $\mathcal{E}(\cdot)$ denotes the keypoint perturbation and sparse-correction simulation process and $\boldsymbol{\xi}$ represents the sampled perturbation.
The decoder then learns to recover the target pose:
\begin{equation}
\hat{\mathbf{K}}^{t}
=
\mathcal{D}
\left(
\mathbf{F}^{t},
\widetilde{\mathbf{K}}^{t}
\right),
\qquad
\hat{\mathbf{K}}^{t}
\rightarrow
\mathbf{K}^{t}.
\end{equation}

The temporal correspondence module is decoupled from this pose-refinement training.
At inference time, temporally propagated human corrections instantiate the same type of imperfect geometric priors seen by the correction-aware decoder.
The learned refinement capability can therefore be applied to propagated corrections without treating the tracking result as a hard target location.

\paragraph{Temporal corrections as geometric priors.}
After the initial pose and identity predictions are obtained, annotators may correct erroneous keypoints, bounding boxes, or identities at frame $t$.
Let the corrected keypoints and boxes be
$\mathbf{K}^{t'}$ and $\mathbf{B}^{t'}$.
As shown in Fig.~\ref{fig: Fig:3.1}(c), we use CoTracker~\cite{karaev24cotracker3} solely as a temporal correspondence operator to estimate candidate locations of the corrected keypoints in subsequent frames:
\begin{equation}
\widetilde{\mathbf{K}}^{t+\Delta t}
=
\mathcal{T}
\left(
\mathbf{K}^{t'}
\right).
\end{equation}

Importantly,
$\widetilde{\mathbf{K}}^{t+\Delta t}$
is not directly used as the target-frame pose.
Instead, the propagated locations are converted into keypoint reference points
$\mathbf{Q}_{K}^{\Delta t}$,
while the corresponding person extent defines box reference points
$\mathbf{Q}_{B}^{\Delta t}$.
Both are normalized to $[0,1]$ image coordinates and mapped into inverse-sigmoid space following the reference-point parameterization of the DETR-style decoders.
The propagated correction therefore specifies where the model should search, rather than prescribing the final pose.

\textbf{Correction-conditioned decoding.}
For a target frame $t+1$, the propagated box reference
$\mathbf{Q}_{B}^{\Delta t}$
conditions the current-frame human content query
$\mathbf{Q}_{C}^{t+1}$
through the Box Decoder:
\begin{equation}
\mathbf{Q}_{B}^{t+1'}
=
D_{B}
\left(
\mathbf{Q}_{C}^{t+1},
\mathbf{Q}_{B}^{\Delta t}
\right),
\end{equation}
where $D_{B}(\cdot)$ denotes the box decoder.

The updated box query and propagated keypoint reference are then passed to the Keypoint Decoder:
\begin{equation}
\left(
\mathbf{Q}_{B}^{t+1''},
\mathbf{Q}_{K}^{t+1'}
\right)
=
D_{K}
\left(
\mathbf{Q}_{B}^{t+1'},
\mathbf{Q}_{K}^{\Delta t},
\mathbf{Q}_{C}^{t+1}
\right),
\end{equation}
where $D_{K}(\cdot)$ denotes the keypoint decoder.
Each decoder layer performs residual refinement around the propagated references.
Consequently, the temporal tracker provides candidate geometric locations, while the decoder determines the final pose from the target-frame observation.

This separation is important because temporal correspondence and pose estimation solve different problems.
Tracking estimates where a corrected point may move, whereas pose decoding determines where the anatomical joint should be given the current image.
By treating tracking outputs as priors rather than labels, IMPose allows the visual evidence of each target frame to revise imperfect temporal correspondence.

\textbf{Closed-loop correction propagation.}
After decoding, the refined pose is associated with the keypoint-aware identity state described in Sec.~\ref{sec:Instance-level Tracking}.
The resulting identity-conditioned tracklet is written back to the trajectory bank and becomes part of the state used for subsequent frames.
The overall refinement process is therefore
\begin{equation}
\mathbf{C}_{t}
\rightarrow
\widetilde{\mathbf{C}}_{t\rightarrow t+\Delta t}
\rightarrow
\hat{\mathbf{P}}_{t+\Delta t}
\rightarrow
\mathbf{S}_{t+\Delta t},
\end{equation}
where
$\mathbf{C}_{t}$ denotes the human correction,
$\widetilde{\mathbf{C}}_{t\rightarrow t+\Delta t}$ its propagated geometric prior,
$\hat{\mathbf{P}}_{t+\Delta t}$ the pose refined from target-frame evidence,
and $\mathbf{S}_{t+\Delta t}$ the updated pose--identity state.

The updated state is then used for later frames.
Thus, a local human correction does not terminate at the edited frame:
it is transformed into a temporal geometric prior,
reconciled with new visual evidence,
associated with a person- and joint-aware state,
and carried forward for subsequent inference.
This feedback process constitutes the closed temporal correction loop of IMPose.

\section{Interactive Annotation Protocol}
\label{sec:annotation_protocol}

\begin{figure}[t!]
  \centering
  \includegraphics[width=\columnwidth]{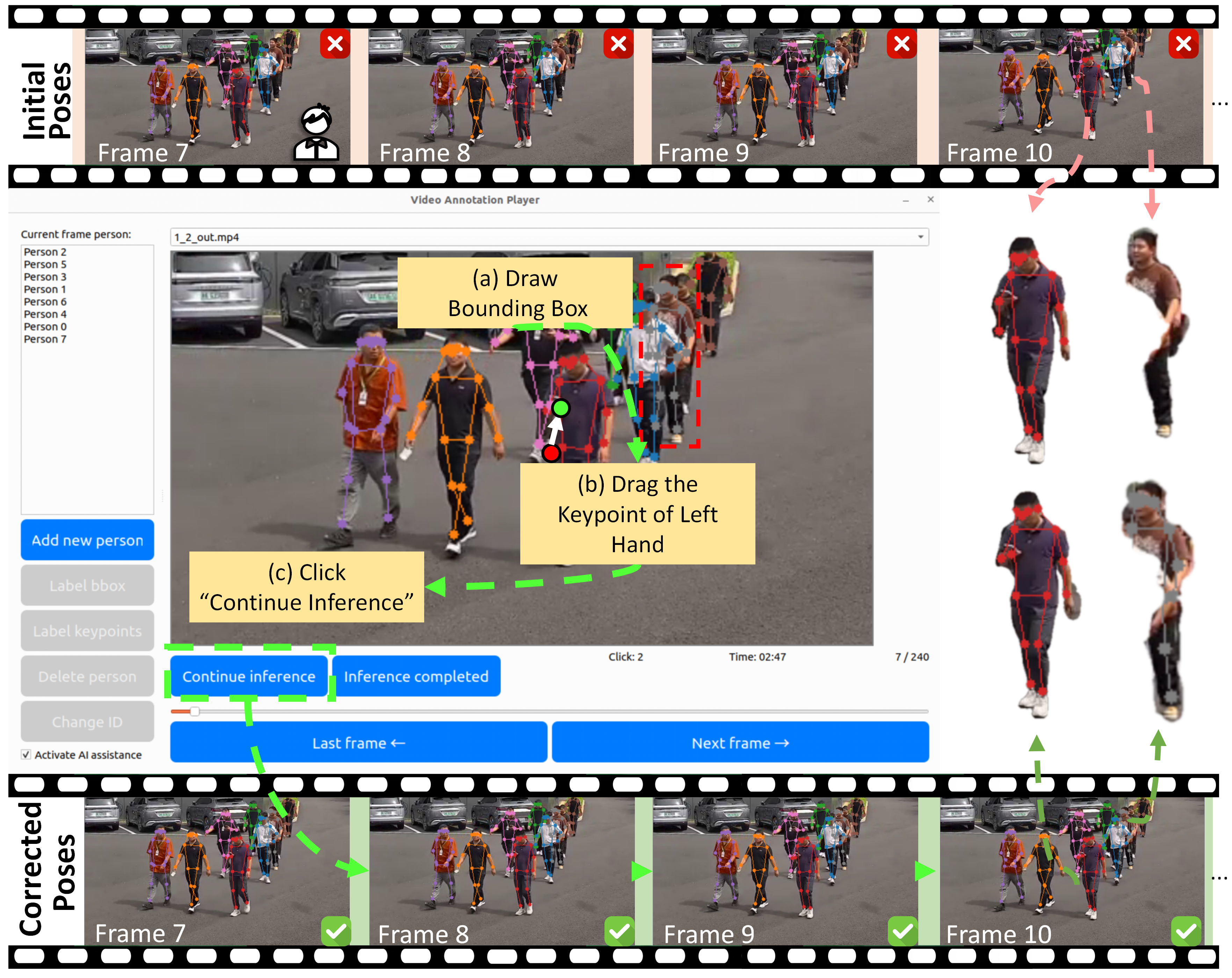}
  \caption{\textbf{Interactive annotation with IMPose.}
  The annotator (a) draws a bounding box to add a missed person, (b) drags a keypoint to correct its location, and (c) clicks \emph{Continue Inference} to propagate the corrections to subsequent frames.
  The upper strip shows the initial predictions, while the lower strip shows the corrected annotations and subsequent predictions.}
  \label{fig:platform}
\end{figure}

To construct dense multi-person pose annotations, we use IMPose within an interactive video annotation workflow. The annotator first runs IMPose to generate initial poses and identities for a video sequence, then corrects erroneous keypoints, bounding boxes, and identity assignments through the annotation interface. Each correction is propagated to subsequent frames by the proposed correction-conditioned inference loop, allowing annotators to verify and refine trajectories rather than relabel each frame independently. Figure~\ref{fig:platform} illustrates this workflow.

We follow a visibility-based filtering protocol to ensure that annotated instances are visually verifiable. For partially occluded or boundary-truncated persons, an instance is annotated only when at least eight joints are clearly visible under the COCO 17-keypoint format. This threshold follows the filtering principle of PoseTrackReID, which removes persons with fewer than six visible joints under a 15-keypoint format, and adapts it to the additional head-related keypoints in COCO. Under severe motion blur, annotators label only joints whose locations remain visually distinguishable; instances with fewer than eight valid joints after this filtering are excluded. Extremely small persons are also omitted when their joint locations cannot be reliably verified.

All annotators receive task-specific training before annotation. The interface supports model-assisted initialization, direct keypoint correction, bounding-box adjustment, and identity addition, deletion, and reassignment. These operations are linked to IMPose, so that a corrected keypoint or identity is not only saved as a local edit but also propagated as a conditioning signal for subsequent frames. This design reduces repeated frame-by-frame correction while preserving annotator control over ambiguous cases. For dataset extension, we count changes to keypoints, bounding boxes, and person IDs together as annotation updates. Additional implementation details and annotation guidelines are provided in Online Resource 1.


\section{Experiments}
\subsection{Experimental Setup}
\begin{figure*}[t]
  \centering
   \includegraphics[width=1\linewidth]{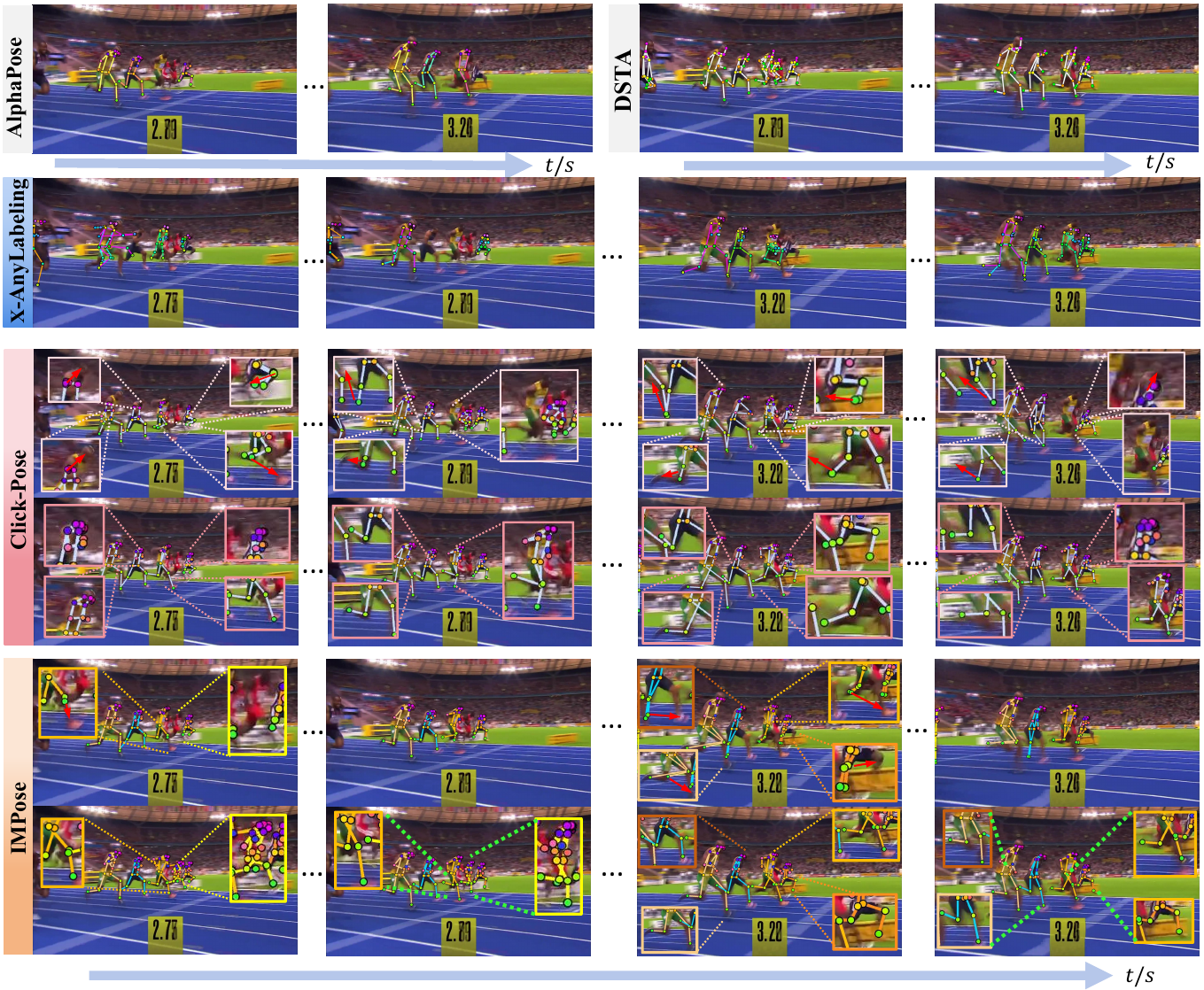}
   \caption{\textbf{Qualitative Comparison between AlphaPose~\cite{alphapose}, DSTA~\cite{he2024decoupled}, Click-Pose~\cite{yang2023neural}, X-AnyLabeling~\cite{wang2023advanced} and IMPose on the In-the-Wild Cases with Body-part Occlusion and Motion Blur}. The green dashed arrow denotes the temporal correction propagation. The upper rows for Click-Pose~\cite{yang2023neural} and IMPose show the correction process, and the lower rows show the results after correction. 
   Notably, poses of DSTA~\cite{he2024decoupled} and Click-Pose~\cite{yang2023neural} are shown in white without cross-frame identity labels in this comparison, while AlphaPose~\cite{alphapose} and X-AnyLabeling~\cite{wang2023advanced} use different colors for each person to represent temporal consistency. This figure demonstrates IMPose produces more stable pose estimations in complex scenarios and maintains better structural accuracy and temporal consistency under motion blur and occlusion. 
   (Zoom in for a better view.) }
   \label{Fig:4.1}
\end{figure*}

\textbf{Datasets.} Drawing on previous experience in pose tracking, we conduct the experiments on the public training sets and validation sets of PoseTrack21~\cite{doering2022posetrack21} and 3DPW~\cite{von2018recovering}. PoseTrack21~\cite{doering2022posetrack21} provides rich annotations for multi-person pose tracking, including occlusions and small objects, while 3DPW~\cite{von2018recovering} serves as a comprehensive outdoor benchmark for pose estimation in diverse social scenarios. 
\textbf{Evaluation Metrics.}
We evaluate annotation effort using Video Number of Clicks (VNoC), defined at a target AP of $\tau\%$ as
\begin{equation}
\mathrm{VNoC}@\tau=\frac{C_\tau}{\displaystyle\sum_{f\in\mathcal F}N_f},
\label{eq:vnoc}
\end{equation}
where $C_\tau$ denotes the cumulative correction clicks required to reach the target AP, $\mathcal F$ is the set of evaluated frames, and $N_f$ is the number of evaluated person instances in frame $f$. Automatic propagation incurs no additional clicks. We report VNoC@75, VNoC@85, and VNoC@95 on the PoseTrack21~\cite{doering2022posetrack21} and 3DPW~\cite{von2018recovering} validation sets, with lower values indicating less interaction effort. We also report overall AP under click constraints to evaluate performance with equal human effort. Object Keypoint Similarity (OKS) assesses temporal keypoint propagation, while Higher Order Tracking Accuracy (HOTA)~\cite{luiten2021hota} balances detection (DetA) and association (AssA) for tracking evaluation, quantifying spatial localization and temporal identity consistency.

\textbf{Comparison methods.}
We compare IMPose with representative methods for automatic pose annotation, interactive pose refinement, and temporal propagation. 
For automatic pose annotation, we include DSTA~\cite{he2024decoupled}, AlphaPose~\cite{alphapose}, and X-AnyLabeling~\cite{wang2023advanced}. DSTA~\cite{he2024decoupled} and AlphaPose~\cite{alphapose} serve as strong automatic baselines for video pose estimation, while X-AnyLabeling~\cite{wang2023advanced} is included as a competitive auto-annotation method based on DWPose. We additionally include DSTA + PoseWarper, which combines DSTA pose estimation with PoseWarper-based temporal propagation.
For interactive pose refinement, we evaluate Click-Pose~\cite{yang2023neural}, which incorporates user corrections within individual images. CoTracker~\cite{karaev24cotracker3} serves as a point-propagation baseline in the correction-propagation experiments. We do not include a full iPose~\cite{liu2024ipose} comparison because its source code is not publicly available. Since iPose uses CoTracker to propagate edited 2D joint locations, we evaluate CoTracker-based point propagation under different correction intervals, motion blur, and occlusion in Sec.~\ref{sec:robustness}. These comparisons assess point propagation without reproducing iPose's subsequent 3D pose and camera optimization.


\subsection{Training}

\textbf{Training Setup.}
During training, we sample $T+1$ frames in each iteration. Following prior multi-frame training strategies~\cite{gao2023memotr,gao2025multiple}, random temporal intervals are adopted to increase training diversity. However, excessively large intervals may introduce overly difficult training samples and create a mismatch with inference-time video sequences. Therefore, we restrict the random sampling interval to the range of 1--4 by default.

Our model adopts a ResNet backbone for feature extraction, which provides strong spatial representations for subsequent pose estimation and tracking modules.

\textbf{Optimization.}
For PoseTrack21~\cite{doering2022posetrack21}, IMPose is trained for 20 epochs on the training set. The learning rate is reduced by a factor of 10 at the 10\textsuperscript{th} and 15\textsuperscript{th} epochs. To accelerate convergence, we initialize the model with COCO~\cite{lin2014microsoft} pre-trained weights and perform keypoint detection pre-training on the corresponding datasets. These weights are used to initialize the Human-to-Keypoint Encoder and Decoder components.

The number of keypoint queries $Q_k$ is set to 50. We employ the AdamW optimizer with an initial learning rate of $1\times10^{-4}$. Training is conducted on 6$\times$A800 GPUs for 20 epochs.

\textbf{Parallelization.}
Recent tracking-by-query methods~\cite{gao2023memotr,yan2023bridging} typically process multi-frame sequences in a sequential manner similar to RNNs. For a $T+1$ frame clip, the DETR module performs $T+1$ forward passes, processing each frame independently. Since the Human-to-Keypoint Decoder accounts for most of the computational cost, such sequential processing limits GPU parallelism.

In contrast, IMPose decouples pose estimation from instance association, allowing keypoint detection to be performed simultaneously across all frames during training. 
Furthermore, the Instance Tracker, implemented as a transformer decoder, introduces additional parallelism through attention masking. As a result, IMPose achieves higher GPU utilization and more efficient training.


\subsection{Inference}
IMPose first generates initial multi-person poses and identity assignments. Annotators then correct erroneous keypoints, bounding boxes, or identities. Corrected keypoints are propagated to subsequent frames, where their estimated locations provide geometric priors for box and keypoint decoding. The decoders refine these priors using target-frame visual evidence, and the resulting poses and identity assignments update the trajectory bank for subsequent inference.

We use a default correction interval of 10 frames. The effect of this interval on propagation accuracy is evaluated in the correction-interval sensitivity experiment.

\subsection{Qualitative Comparison}
We show an in-the-wild video that contains body-part occlusion and motion blur for qualitative comparison in Figure~\ref{Fig:4.1}. AlphaPose~\cite{alphapose}, DSTA~\cite{he2024decoupled} and X-AnyLabeling~\cite{wang2023advanced} are automated annotation methods without spatio-temporal error correction, while DSTA~\cite{he2024decoupled} and Click-Pose~\cite{yang2023neural} are evaluated without cross-frame identity association in this comparison.
Although Click-Pose~\cite{yang2023neural} enables spatial keypoint refinement (requiring about 5 manual clicks per person in Figure~\ref{Fig:4.1} annotation), it relies on frame-by-frame corrections while needing extra clicks to link persons between two frames. In contrast, our IMPose framework achieves dynamic pose annotation across video segments with an average of 2 manual clicks per person, leveraging spatio-temporal correction propagation to maintain annotation consistency. Particularly, once a person fails to be detected, for example the red athlete in the first column of Figure~\ref{Fig:4.1}, an annotator only needs to draw a bounding box around him and then the pose and the poses in the following frames are automatically corrected.

\subsection{Quantitative Comparison}
\label{sec: quan}

\textbf{Video NoC Comparison.} VNoC (Video Number of Clicks) measures the manual corrections required for accurate keypoint annotations. 
As shown in Table~\ref{tab: quantatitive-fixed-AP}, IMPose consistently outperforms other methods across all interaction levels. At VNoC@95, IMPose reduces manual effort by 3.11 and 4.14 corrections on the 3DPW~\cite{von2018recovering} and PoseTrack21~\cite{doering2022posetrack21} datasets, respectively, compared to AlphaPose~\cite{alphapose}. Compared with DSTA + PoseWarper, IMPose reduces VNoC@95 by 1.86 and 0.85 on 3DPW and PoseTrack21, respectively. Moreover, under zero-click conditions, IMPose achieves comparable pose accuracy to X-AnyLabeling (38.87 vs. 38.84 mAP), while obtaining higher HOTA and DetA.
These results highlight the effectiveness of IMPose's correction propagation mechanism, which transforms sparse user inputs into accurate, temporally consistent keypoint annotations with minimal human intervention. IMPose incorporates propagated corrections into subsequent pose decoding as geometric priors and carries the refined pose and identity state forward in time.

\begin{table}[t]
  \centering
  \imposetablestyle
  \caption{Comparisons of Video Number of Clicks (VNoC) for Interactive Keypoint Annotation. IMPose achieves the accuracies with minimal clicks compared to other methods. Dashed lines separate automatic annotation baselines from methods with correction refinement or propagation mechanisms.}
  \label{tab: quantatitive-fixed-AP}
  \setlength{\tabcolsep}{3pt}
  \begin{tabularx}{\columnwidth}{c|l|*{3}{>{\raggedleft\arraybackslash}X}}
    \toprule
    & \multirow{2}{*}{Method} & \multicolumn{3}{c}{VNoC$\downarrow$} \\
    \cmidrule(lr){3-5}
    & & @75 & @85 & @95 \\
    \midrule
    \multirow{6}{*}{\rotatebox[origin=c]{90}{\textit{3DPW}}}
& AlphaPose~\cite{alphapose} & 2.23 & 2.36 & 3.15 \\
& DSTA~\cite{he2024decoupled} & 1.16 & 1.87 & 2.82 \\
& X-AnyLabeling~\cite{wang2023advanced} & 0.97 & 1.14 & 1.99 \\[2pt]
\cdashline{2-5}[2pt/2pt]
& \tablegroupstrut Click-Pose~\cite{yang2023neural} & 1.06 & 1.43 & 2.43 \\
& DSTA + PoseWarper & 0.49 & 1.10 & 1.90 \\
& \cellcolor{imposerow}\textbf{IMPose} & \cellcolor{imposerow}\textbf{0.026} & \cellcolor{imposerow}\textbf{0.026} & \cellcolor{imposerow}\textbf{0.037} \\
\midrule
    \multirow{6}{*}{\rotatebox[origin=c]{90}{\textit{PoseTrack21}}}
& AlphaPose~\cite{alphapose} & 7.66 & 7.92 & 8.69 \\
& DSTA~\cite{he2024decoupled} & 4.37 & 4.95 & 5.53 \\
& X-AnyLabeling~\cite{wang2023advanced} & 4.00 & 4.38 & 5.03 \\[2pt]
\cdashline{2-5}[2pt/2pt]
& \tablegroupstrut Click-Pose~\cite{yang2023neural} & 4.21 & 4.68 & 5.61 \\
& DSTA + PoseWarper & 4.20 & 4.78 & 5.40 \\
& \cellcolor{imposerow}\textbf{IMPose} & \cellcolor{imposerow}\textbf{3.39} & \cellcolor{imposerow}\textbf{4.37} & \cellcolor{imposerow}\textbf{4.55} \\
\bottomrule
  \end{tabularx}
\end{table}

\begin{table*}[t]
  \centering
  \imposetablestyle
  \scriptsize
  \caption{\textbf{Quantitative Comparisons under Varying Simulated Correction Budgets.}
    AlphaPose~\cite{alphapose}, DSTA~\cite{he2024decoupled}, and X-AnyLabeling~\cite{wang2023advanced} provide automatic annotation baselines. DSTA + PoseWarper combines pose estimation with temporal propagation, while Click-Pose~\cite{yang2023neural} and IMPose support interactive pose refinement. Corrections are simulated by replacing selected keypoints with ground-truth annotations. Dashed lines separate automatic annotation baselines from methods with correction refinement or propagation mechanisms. IMPose achieves the highest overall mAP, HOTA, and DetA at each evaluated correction budget.}

  \label{tab:quantative-fixed-clicks}

  \setlength{\tabcolsep}{1.5pt}
  
  \begin{adjustbox}{max width=\textwidth}
  \begin{tabular}{c|l|rr|*{18}{r}}
    \toprule
    \multirow{2}{*}{Clicks} & \multirow{2}{*}{Method} 
    & \multicolumn{2}{c|}{Tracking metrics} 
    & \multicolumn{18}{c}{Pose AP metrics} \\
    \cmidrule(lr){3-4} \cmidrule(lr){5-22}
    & 
    & \scriptsize{HOTA$\uparrow$} & \scriptsize{DetA$\uparrow$}
    & \scriptsize{mAP$\uparrow$}
    & \scriptsize{Nose} & \scriptsize{L-Eye} & \scriptsize{R-Eye} & \scriptsize{L-Ear} & \scriptsize{R-Ear}
    & \scriptsize{L-Sho} & \scriptsize{R-Sho} & \scriptsize{L-Elb} & \scriptsize{R-Elb} & \scriptsize{L-Wri} & \scriptsize{R-Wri}
    & \scriptsize{L-Hip} & \scriptsize{R-Hip} & \scriptsize{L-Kne} & \scriptsize{R-Kne} & \scriptsize{L-Ank} & \scriptsize{R-Ank} \\
    \midrule

\multirow{6}{*}{0}
& AlphaPose & 40.89 & 33.40 & 28.90 
& 20.33 & - & - & - & -
& 35.73 & 35.36 & 26.24 & 25.29 & 15.23 & 15.17
& 35.05 & 34.95 & 28.48 & 28.09 & 24.53 & 24.21 \\
& DSTA & - & - & 29.63
& 28.53  & - & - & - & -
& \textbf{51.83} & 51.51 & 38.55 & 39.31 & 30.14 & 31.48
& 47.96 & 48.39 & 35.64 & 35.07 & 35.20 & 34.32 \\
& X-AnyLabeling & 46.70 & 36.13 & 38.84
& 27.05 & 26.43 & 26.37 & 33.61 & 33.70
& 51.64 & \textbf{52.01} & 42.65 & 42.52 & 32.71 & 32.38
& \textbf{55.30} & \textbf{55.41} & 43.47 & 42.94 & 39.62 & 38.49 \\[2pt]
\cdashline{2-22}[2pt/2pt]
& \tablegroupstrut Click-Pose & - & - & 30.51
& 28.15 & 28.89 & 29.14 & 28.84 & 28.95
& 33.39 & 33.26 & 28.17 & 27.93 & 20.99 & 20.92
& 37.81 & 35.84 & 26.91 & 24.28 & 22.47 & 19.54 \\
& DSTA + PoseWarper & - & - & 33.73
& \textbf{37.85} & - & - & - & -
& 47.94 & 48.12 & \textbf{44.40} & \textbf{45.20} & \textbf{37.74} & \textbf{39.82}
& 48.74 & 48.58 & \textbf{44.85} & \textbf{44.40} & \textbf{41.06} & \textbf{40.73} \\
\rowcolor{imposerow}
& \textbf{IMPose} & \textbf{52.21} & \textbf{42.91} & \textbf{38.87}
& 37.39 & \textbf{37.34} & \textbf{37.00} & \textbf{40.04} & \textbf{40.08}
& 45.97 & 45.87 & 41.44 & 40.46 & 32.96 & 32.52
& 45.56 & 44.98 & 37.23 & 36.46 & 33.82 & 34.97 \\
\midrule

\multirow{6}{*}{1}
& AlphaPose & 40.89 & 33.40 & 31.74
& 23.93 & - & - & - & -
& 35.79 & 35.44 & 28.49 & 27.50 & 26.56 & 26.69
& 35.10 & 35.06 & 30.17 & 29.58 & 28.19 & 28.41 \\
& DSTA & - & - & 37.82
& 33.82 & - & - & - & -
& 52.16 & 51.95 & 39.59 & 40.40 & 35.93 & 37.30 
& 48.25 & 48.68 & 36.47 & 35.79 & 37.06 & 36.65 \\
& X-AnyLabeling & 46.70 & 36.13 & 44.35
& 34.66 & 39.02 & 37.44 & 35.27 & 35.31
& 52.12 & 52.43 & 44.42 & 44.16 & 43.28 & 43.45
& 55.67 & 55.80 & 45.70 & 45.58 & 45.51 & 45.49 \\[2pt]
\cdashline{2-22}[2pt/2pt]
& \tablegroupstrut Click-Pose & - & - & 35.56
& 37.05 & 37.80 & 37.47 & 32.63 & 32.89
& 36.43 & 36.28 & 29.82 & 29.92 & 36.26 & 35.61
& 38.29 & 36.45 & 29.08 & 27.88 & 27.71 & 28.21 \\
& DSTA + PoseWarper & - & - & 41.50
& 38.83 & - & - & - & -
& 47.94 & 48.12 & 44.40 & 45.20 & 37.74 & 39.82
& 48.74 & 48.58 & 44.85 & 44.40 & 41.05 & 40.85 \\
\rowcolor{imposerow}
& \textbf{IMPose} & \textbf{64.89} & \textbf{63.41} & \textbf{67.99}
& \textbf{59.10} & \textbf{58.76} & \textbf{58.28} & \textbf{61.95} & \textbf{62.90}
& \textbf{73.36} & \textbf{73.08} & \textbf{66.46} & \textbf{66.75} & \textbf{63.82} & \textbf{63.49}
& \textbf{74.25} & \textbf{74.79} & \textbf{63.45} & \textbf{62.92} & \textbf{63.44} & \textbf{63.04} \\
\midrule

\multirow{6}{*}{2}
& AlphaPose & 40.89 & 33.40 & 33.92
& 28.83 & - & - & - & -
& 35.98 & 35.60 & 32.27 & 31.54 & 31.71 & 31.73
& 35.22 & 35.23 & 31.97 & 31.80 & 32.01 & 31.97 \\
& DSTA & - & - & 42.21
& 36.41 & - & - & - & -
& 53.12 & 52.80 & 40.72 & 41.87 & 38.97 & 40.42 
& 48.56 & 49.01 & 37.35 & 36.71 & 39.23 & 38.65 \\
& X-AnyLabeling & 46.70 & 36.13 & 48.09
& 37.98 & 43.54 & 42.21 & 37.35 & 37.21
& 52.44 & 52.74 & 45.33 & 45.02 & 45.38 & 45.57
& 55.90 & 56.10 & 46.68 & 46.57 & 47.57 & 47.51 \\[2pt]
\cdashline{2-22}[2pt/2pt]
& \tablegroupstrut Click-Pose & - & - & 39.04
& 40.74 & 41.32 & 41.08 & 36.20 & 36.70
& 37.30 & 37.34 & 32.29 & 32.66 & 37.94 & 37.64
& 38.59 & 36.89 & 29.85 & 28.64 & 30.49 & 30.65 \\
& DSTA + PoseWarper & - & - & 44.98
& 38.98 & - & - & - & -
& 48.08 & 48.11 & 44.40 & 45.20 & 37.88 & 39.95
& 48.74 & 48.58 & 44.84 & 44.41 & 41.04 & 40.71 \\
\rowcolor{imposerow}
& \textbf{IMPose} & \textbf{66.71} & \textbf{66.10} & \textbf{75.10}
& \textbf{64.04} & \textbf{63.77} & \textbf{63.25} & \textbf{66.19} & \textbf{67.33}
& \textbf{77.61} & \textbf{77.60} & \textbf{73.47} & \textbf{74.25} & \textbf{73.00} & \textbf{72.93}
& \textbf{79.11} & \textbf{79.39} & \textbf{71.53} & \textbf{71.35} & \textbf{74.51} & \textbf{73.66} \\
\midrule

\multirow{6}{*}{3}
& AlphaPose & 40.89 & 33.40 & 35.50
& 33.99 & - & - & - & -
& 36.32 & 35.93 & 34.39 & 33.95 & 34.57 & 34.54
& 35.55 & 35.52 & 34.12 & 33.96 & 34.60 & 34.42 \\
& DSTA & - & - & 54.20
& 55.17 & - & - & - & -
& 57.60 & 57.81 & 51.64 & 52.40 & 58.07 & 58.54
& 51.97 & 52.24 & 47.59 & 47.11 & 52.77 & 53.07 \\
& X-AnyLabeling & 46.70 & 36.13 & 56.42
& 47.19 & 51.76 & 50.94 & 40.40 & 40.36
& 53.00 & 53.22 & 47.25 & 47.13 & 49.55 & 49.83
& 56.33 & 56.64 & 49.15 & 49.41 & 52.13 & 52.04 \\[2pt]
\cdashline{2-22}[2pt/2pt]
& \tablegroupstrut Click-Pose & - & - & 47.12
& 46.43 & 46.58 & 46.58 & 38.82 & 41.22
& 39.13 & 39.09 & 37.35 & 38.09 & 45.81 & 43.50
& 39.24 & 37.72 & 35.91 & 32.93 & 37.53 & 36.52 \\
& DSTA + PoseWarper & - & - & 56.68
& 57.32 & - & - & - & -
& 58.24 & 58.27 & 54.52 & 55.19 & 47.88 & 50.08
& 58.73 & 58.57 & 54.85 & 54.40 & 51.04 & 50.83 \\
\rowcolor{imposerow}
& \textbf{IMPose} & \textbf{67.25} & \textbf{67.30} & \textbf{76.15}
& \textbf{67.39} & \textbf{67.35} & \textbf{66.81} & \textbf{69.34} & \textbf{70.47}
& \textbf{80.24} & \textbf{80.37} & \textbf{79.47} & \textbf{78.83} & \textbf{78.21} & \textbf{78.46}
& \textbf{82.08} & \textbf{82.43} & \textbf{78.61} & \textbf{77.32} & \textbf{80.26} & \textbf{82.31} \\
\bottomrule
\end{tabular}
  \end{adjustbox}
\end{table*}

\textbf{Interactive Temporal Pose Annotation.}
To evaluate IMPose for interactive temporal pose annotation, we compare it with AlphaPose~\cite{alphapose}, DSTA~\cite{he2024decoupled}, DSTA + PoseWarper, Click-Pose~\cite{yang2023neural}, and X-AnyLabeling~\cite{wang2023advanced} under controlled user interactions ranging from 0 to 3 clicks. Following a standardized simulation protocol, we first compute the Object Keypoint Similarity (OKS) error between each predicted keypoint and its ground-truth annotation. For each method, we then identify the top-$C$ keypoints with the largest OKS errors and directly replace them with ground-truth annotations to simulate $C$ manual corrections, where $C \in \{0,1,2,3\}$. The case of $C=0$ corresponds to fully automatic annotation without human intervention. This controlled protocol enables us to quantitatively evaluate how effectively different methods can exploit sparse user feedback. Table~\ref{tab:quantative-fixed-clicks} reports overall pose accuracy (mAP), per-keypoint AP, and tracking quality (HOTA and DetA), allowing a detailed analysis of how annotation quality evolves as the interaction budget increases.

As shown in Table~\ref{tab:quantative-fixed-clicks}, IMPose achieves the best performance--interaction trade-off across all click budgets, and its advantage becomes substantially more pronounced once sparse user corrections are introduced. In the 0-click setting, which mainly reflects automatic initialization quality, IMPose remains comparable with the strongest automatic baseline in pose accuracy while achieving the best tracking performance, with HOTA/DetA improving from 46.70/36.13 for X-AnyLabeling~\cite{wang2023advanced} to 52.21/42.91. This indicates that IMPose not only provides a strong initialization for subsequent refinement, but also produces more temporally stable and identity-consistent pose trajectories before user intervention. 
With similar automatic initialization accuracy, the larger gains after correction indicate that IMPose makes more effective use of sparse feedback under the evaluated protocol.

More importantly, the superiority of IMPose becomes much clearer in the low-click regime, which is the practically relevant setting for interactive annotation. With only one user correction, IMPose improves mAP from 38.87 to 67.99, yielding a gain of 29.12 points, whereas X-AnyLabeling improves from 38.84 to 44.35, a gain of only 5.51 points. In comparison, DSTA + PoseWarper improves from 33.73 to 41.50 mAP, a gain of 7.77 points. Additionally, the tracking quality of IMPose also improves dramatically, with HOTA/DetA increasing from 52.21/42.91 to 64.89/63.41, while X-AnyLabeling remains unchanged at 46.70/36.13. Even with two clicks, IMPose further reaches 75.10 mAP, substantially outperforming X-AnyLabeling at 48.09 and DSTA at 42.21, while also achieving 66.71 HOTA and 66.10 DetA. This trend persists as the click budget increases: at 3 clicks, IMPose reaches 76.15 mAP, 67.25 HOTA, and 67.30 DetA, outperforming the strongest pose baseline, DSTA + PoseWarper (56.68 mAP), by 19.47 mAP points. These results show that IMPose is not merely competitive in automatic initialization, but significantly more effective at converting sparse human feedback into large gains in both pose accuracy and temporal tracking quality.

Notably, once sparse corrections are introduced, the advantage of IMPose rapidly extends to nearly all body joints. The improvements are particularly evident on challenging extremities such as wrists, knees, and ankles, which are typically more vulnerable to occlusion, motion blur, and temporal inconsistency. For example, under one click, IMPose already achieves 63.82/63.49 AP on the wrists and 63.44/63.04 AP on the ankles, substantially exceeding the corresponding scores of X-AnyLabeling 43.28/43.45 for wrists and 45.51/45.49 for ankles. Similar margins remain under two and three clicks, indicating that sparse corrections are not only absorbed locally, but are effectively propagated to difficult joints over time.

These results show that sparse corrections improve pose accuracy and tracking quality within the correction-conditioned inference framework. Propagated corrections provide geometric priors for target-frame visual refinement, and the refined pose and identity state are written back to the trajectory bank for subsequent inference. This connects local human feedback with pose estimation and identity association across frames.

\begin{figure}[t]
  \centering
  \includegraphics[width=\columnwidth]{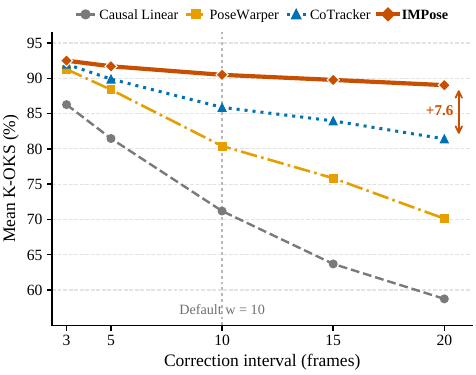}
  \caption{\textbf{Sensitivity of Single-joint Correction Propagation to Different Correction Intervals.} Causal Linear, PoseWarper~\cite{posewarper}, CoTracker~\cite{karaev24cotracker3}, and IMPose use identical IMPose initialization, ground-truth correction anchors, person trajectories, and evaluation ranges. Results are averaged equally over five selected PoseTrack21~\cite{doering2022posetrack21} validation sequences. IMPose maintains consistently higher K-OKS as the correction interval increases.}
  \label{fig:interval_sensitivity}
\end{figure}

\subsection{Robustness Analysis}
\label{sec:robustness}

\begin{figure*}[t]
\centering
\includegraphics[width=\linewidth]{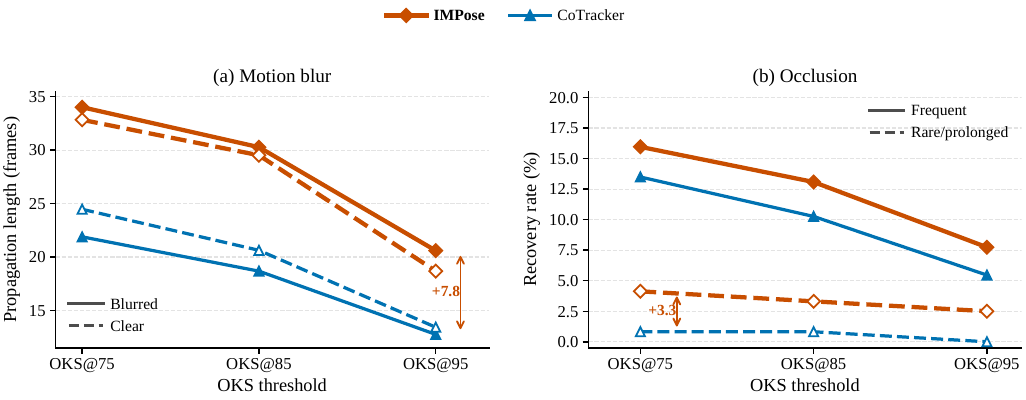}
\caption{\textbf{Correction Propagation Comparison of IMPose and CoTracker~\cite{karaev24cotracker3} under Motion Blur (Left) and Occlusion (Right).} The left line chart displays the average number of frames wherein the corrected keypoints maintain OKS thresholds for blurred and clear videos. The right line chart illustrates the successfully propagated proportion of keypoints after occlusion under frequent and rare settings. Across all conditions, IMPose consistently demonstrates superior corrected keypoint propagation robustness compared to CoTracker.}
\label{fig:occlu_blur}
\end{figure*}

We compare the capabilities of IMPose against CoTracker~\cite{karaev24cotracker3} under motion blur and occlusion. 
iPose~\cite{liu2024ipose} uses CoTracker to propagate edited 2D joint locations. As iPose's source code is not publicly available, we use CoTracker~\cite{karaev24cotracker3} as the point-propagation baseline in Figs.~\ref{fig:interval_sensitivity} and~\ref{fig:occlu_blur}. We also examine how IMPose propagates manual corrections to recover from person identity mismatches.
All experiments are conducted on the PoseTrack21~\cite{doering2022posetrack21} validation set.

\textbf{Sensitivity to Correction Interval.}
To investigate how the temporal distance between consecutive user corrections affects propagation accuracy, we evaluate Causal Linear extrapolation, PoseWarper~\cite{posewarper}, CoTracker~\cite{karaev24cotracker3}, and IMPose using correction intervals of 3, 5, 10, 15, and 20 frames. To isolate correction propagation from automatic pose initialization, all methods start from the same IMPose predictions and receive identical ground-truth corrections for the same persons, keypoints, and temporal ranges. At each correction frame, we traverse every visible joint of every person in turn, replace the selected joint with its ground-truth location to simulate a single manual correction, and propagate it to the subsequent frames. We report mean keypoint-level OKS (K-OKS) over five selected PoseTrack21~\cite{doering2022posetrack21} validation sequences, with each sequence contributing equally to the final average.

As shown in Fig.~\ref{fig:interval_sensitivity}, propagation accuracy decreases with longer correction intervals for all methods, while IMPose exhibits the smallest degradation. Its mean K-OKS decreases from 92.50\% at an interval of 3 frames to 89.03\% at 20 frames, corresponding to a drop of only 3.47 points. At the longest interval, IMPose still outperforms CoTracker, the strongest competing propagation baseline, by 7.62 points (89.03\% vs. 81.41\%). In comparison, PoseWarper and Causal Linear decrease to 70.12\% and 58.74\%, respectively, showing that local warping or linear motion extrapolation is more susceptible to accumulated temporal errors.

At the default interval of 10 frames, IMPose achieves 90.49\% K-OKS. Compared with the 5-frame setting, doubling the interval halves the correction frequency while reducing K-OKS by only 1.21 points. We therefore adopt 10 frames as the default setting, providing a favorable trade-off between propagation accuracy and user interaction frequency. These results demonstrate that the proposed correction propagation mechanism preserves sparse single-joint corrections over longer temporal ranges and reduces the need for frequent manual intervention.

\begin{figure*}[t!]
  \centering
   \includegraphics[width=1\linewidth]{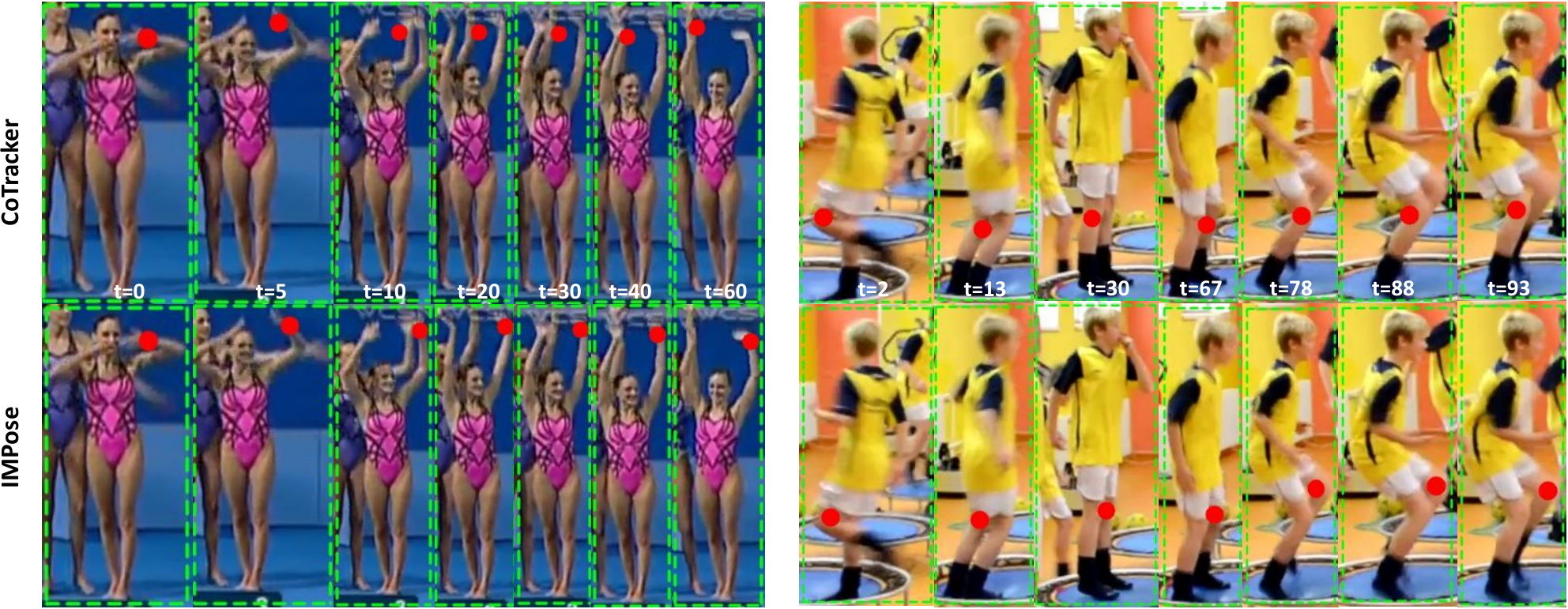}
   \caption{\textbf{Comparative Capability of Corrected Keypoint Propagation between CoTracker~\cite{karaev24cotracker3} and IMPose under Motion Blur.} The corrected keypoint (denoted as red dot) in the first frame of each case is the manually annotated keypoint. The left case is to propagate the left wrist, while the right case is to propagate the left knee.}
   \label{Fig:motion_blur}
\end{figure*}
\textbf{Robustness to Motion Blur.}
To evaluate the robustness of IMPose under motion blur, we construct a challenging evaluation subset based on blur statistics. Specifically, we compute blur scores for all video frames using a Laplacian operator. Videos are then divided into two groups according to these scores: the top one-third with the highest blur scores are categorized as blurred videos, while the bottom one-third with the lowest scores are regarded as clear videos. 

For each video, we select the frame with the highest blur score within the first half of the sequence as the annotation frame, whose ground-truth poses are treated as the initial corrected keypoints. We then compare IMPose with CoTracker~\cite{karaev24cotracker3} by measuring the number of frames in which all propagated keypoints remain correct, where correctness is defined by OKS thresholds of 0.75, 0.85, and 0.95.


As shown in the left part of Fig.~\ref{fig:occlu_blur}, IMPose maintains predefined OKS thresholds for a higher average number of frames across both blurred and clear videos. Under motion blur at OKS@75, IMPose maintains corrected-keypoint accuracy for an average of 34.02 frames, compared with 21.89 frames for CoTracker. At the stricter OKS@95 threshold, IMPose attains 20.60 frames compared with 12.76 frames for CoTracker~\cite{karaev24cotracker3}. IMPose also attains a slightly higher average on blurred videos than on clear videos at OKS@75 (34.02 vs. 32.84 frames). Because these groups also differ in scene dynamics and occlusion, this between-group comparison does not isolate the effect of motion blur.


\begin{figure*}[t!]
  \centering
   \includegraphics[width=1\linewidth]{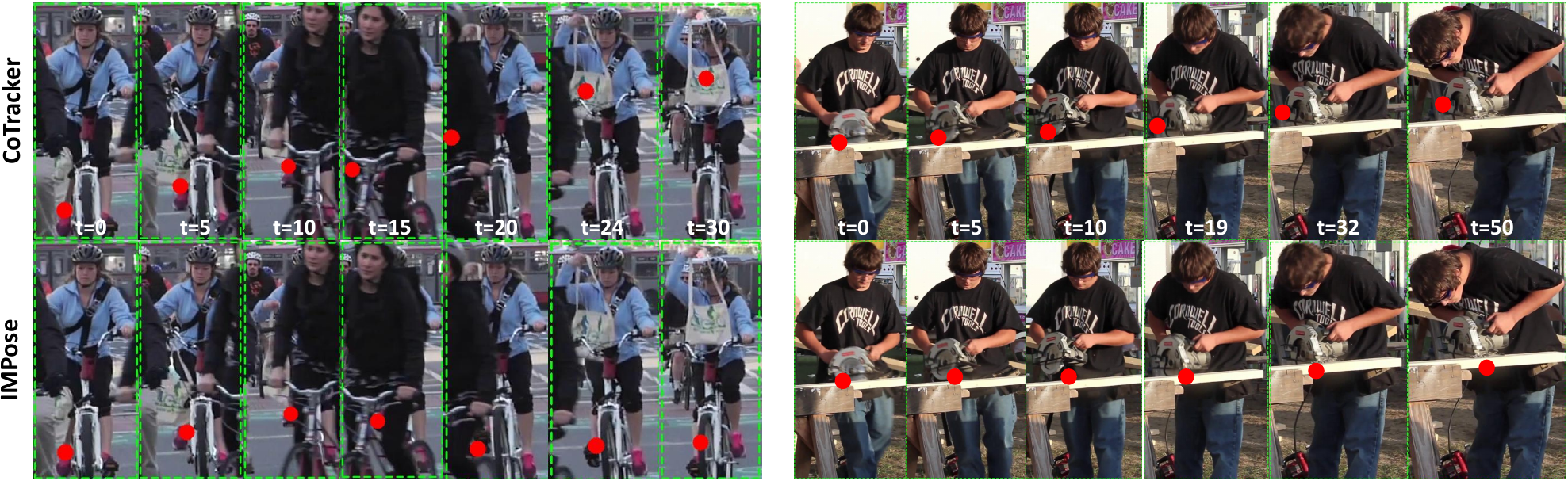}
   \caption{\textbf{Comparative Capability of Corrected Keypoint Propagation between CoTracker~\cite{karaev24cotracker3} and IMPose under Occlusion.}The corrected keypoint (denoted as red dot) in the first frame of each case is the manually annotated keypoint. The left case is to propagate the right ankle of the lady in blue, while the right case is to propagate the right hip of the man wearing jeans. }
   \label{Fig:occlusion}
\end{figure*}

Moreover, Fig.~\ref{Fig:motion_blur} provides two representative examples from videos with severe motion blur.
In the first case, during the waving motion of the artistic swimmer, CoTracker~\cite{karaev24cotracker3} fails to accurately track the left wrist due to the strong blur caused by rapid arm movement. In contrast, IMPose maintains stable and precise tracking of this keypoint throughout the entire sequence. 
In the second example, involving a boy performing a high kick, CoTracker~\cite{karaev24cotracker3} loses the left knee keypoint during the fast upward motion. By comparison, IMPose consistently preserves the correct keypoint trajectory without interruption. 
In IMPose, propagated keypoints provide geometric priors for the correction-aware pose decoder, which refines their locations using target-frame visual evidence. The refined poses are then associated with a keypoint-aware identity state and used to update the trajectory bank. The examples illustrate the behavior of this correction-conditioned inference process under motion blur.

\begin{figure*}[t!]
  \centering
   \includegraphics[width=1\linewidth]{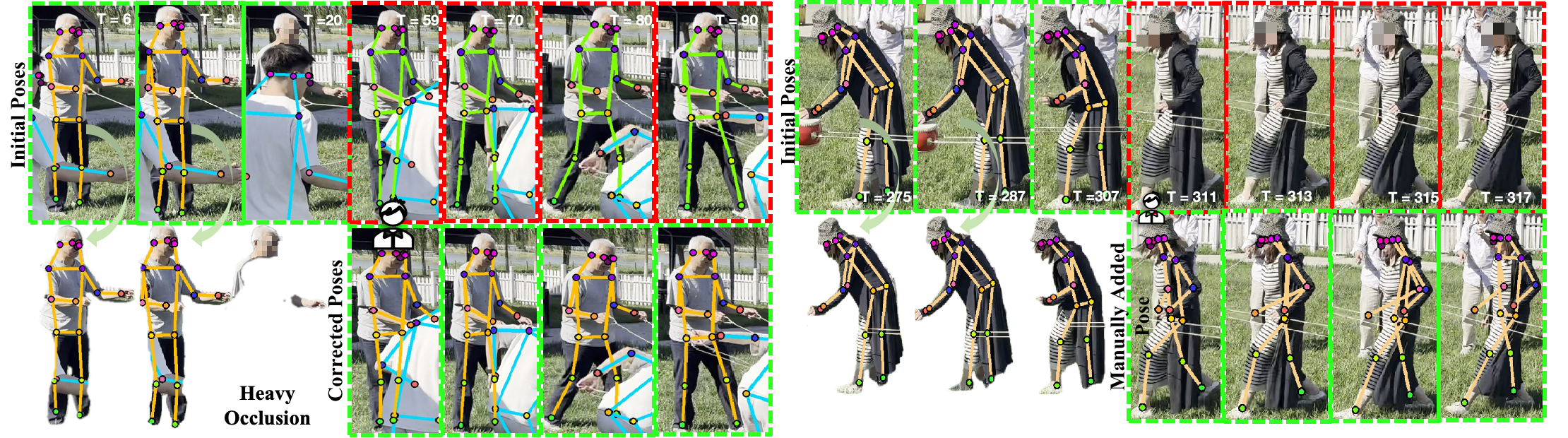}
   \caption{\textbf{The Capability of Corrected Keypoint Propagation of IMPose under ID Mismatching.} The first row of each case presents the initial poses, while the second row of each case presents the corrected poses via IMPose. The frame covered by a cartoon man means the manual correction appears. The following frames without the cartoon man but with green bounding boxes denote the instance id and poses are automatically updated via IMPose. The top shows that IMPose propagates the correction of the misdetected instance to the following frames, while the bottom illustrates the propagation of shifting ID to the man.}
   \label{Fig:id_wrong}
\end{figure*}

\textbf{Robustness to Occlusion.} 
To evaluate robustness under occlusion, we construct test samples from our newly annotated ground-truth data. For each occluded keypoint, we measure the occlusion duration as the temporal interval between its disappearance and reappearance. Based on the frequency distribution of these durations, video segments are categorized into two groups: \emph{frequently occluded} average occlusion length: 16.5 frames and \emph{rare but prolonged occlusion} average occlusion length: 64.8 frames. 

To simulate interactive correction, each occluded keypoint is initialized with its ground-truth location from the last visible frame, and the corrected point is propagated forward using IMPose and CoTracker. Recovery success is determined by evaluating the Object Keypoint Similarity (OKS) when the keypoint reappears, using predefined OKS thresholds.

As shown in the right part of Fig.~\ref{fig:occlu_blur}, IMPose consistently achieves higher recovery rates across all occlusion settings and OKS thresholds. For example, under frequent occlusion at OKS@75, IMPose achieves 15.96\%, outperforming CoTracker 13.49\%. The advantage becomes more pronounced under prolonged occlusion, where IMPose attains 4.13\% compared with only 0.83\% achieved by CoTracker~\cite{karaev24cotracker3}. These results show higher post-occlusion keypoint recovery rates for IMPose, although recovery after prolonged occlusion remains low.

We further present two qualitative examples in Fig.~\ref{Fig:occlusion}. In each video, a keypoint is manually annotated before occlusion and then propagated to subsequent frames using both methods. 

As shown in the left example, when the right ankle of the lady in blue becomes occluded, CoTracker fails to maintain a stable trajectory for this keypoint. In contrast, IMPose successfully preserves the correct keypoint position throughout the occlusion period. In the right example, CoTracker produces severe tracking drift by incorrectly associating the right hip with a nearby wood-cutting machine. By comparison, IMPose maintains accurate localization of the right hip despite the heavy occlusion caused by the equipment.

IMPose refines propagated locations using target-frame visual evidence and associates the resulting poses with a keypoint-aware identity state. This provides a mechanism for revising inaccurate temporal correspondences while retaining person- and joint-level context.

\textbf{Robustness to ID Mismatching.} 
Instance identity mismatching typically arises from two scenarios: (1) missed detections that lead to track interruption, and (2) incorrect identity assignment between nearby instances. We illustrate representative cases in Fig.~\ref{Fig:id_wrong} using an in-the-wild video sequence.

The results show that once a manual correction either to the identity label or the keypoint pose is introduced, IMPose can effectively propagate the correction to subsequent frames, restoring temporally consistent pose predictions. In the top example of Fig.~\ref{Fig:id_wrong}, a previously undetected person is successfully recovered after a single correction. In the bottom example, an identity swap between two individuals is corrected and remains stable throughout the remaining frames.

These results demonstrate that the proposed correction propagation mechanism enables IMPose to effectively recover from identity mismatches and maintain coherent instance-level tracking over time.



\subsection{Ablation Study}
\begin{table}[tb]
  \centering
  \imposetablestyle
  \caption{Ablation of the keypoint-aware pose embedding (KP-AE) for person instance tracking and temporal point tracking for correction propagation. Combining both components yields the highest mAP, HOTA, and AssA and the lowest VNoC@95 among the evaluated variants.}
  \label{tab:4}
  \setlength{\tabcolsep}{2pt}
  \ADLinactivate
  \begin{tabularx}{\columnwidth}{cc|*{3}{>{\raggedleft\arraybackslash}X}|r}
    \toprule
    KP-AE & \shortstack{Point\\tracking} & mAP$\uparrow$ & HOTA$\uparrow$ & AssA$\uparrow$ & VNoC@95$\downarrow$ \\
    \midrule
    \XSolidBrush & \XSolidBrush & 45.73 & 42.80 & 39.27 & 12.06 \\
    \Checkmark & \XSolidBrush & 47.88 & 53.27 & 57.30 & 11.97 \\
    \XSolidBrush & \Checkmark & 72.40 & 61.23 & 57.12 & 7.63 \\
    \rowcolor{imposerow}
    \Checkmark & \Checkmark & \textbf{76.15} & \textbf{67.25} & \textbf{68.50} & \textbf{4.55} \\
    \bottomrule
  \end{tabularx}
\end{table}
We examine how temporal geometric priors and pose-aware instance tracking contribute to correction-conditioned inference. Disabling KP-AE tests the contribution of structured pose geometry to tracking person instances and maintaining their identities across frames. Disabling point tracking tests the contribution of cross-frame correspondence to supplying geometric priors for subsequent pose refinement. Pose and tracking metrics are evaluated under three corrections per person.

As shown in Table~\ref{tab:4}, with point tracking enabled, removing KP-AE reduces HOTA from 67.25 to 61.23 and AssA from 68.50 to 57.12, demonstrating the importance of pose geometry for person instance tracking during correction propagation. With KP-AE enabled, disabling point tracking reduces mAP from 76.15 to 47.88, while VNoC@95 increases from 4.55 to 11.97. This highlights the role of temporal correspondence in extending the influence of sparse corrections beyond the edited frame. Combining both components achieves the best results across all reported metrics, demonstrating the benefit of coupling temporal geometric priors with pose-aware instance tracking within the closed temporal correction loop.

\subsection{Time Cost under Real Manual Corrections}

In Table~\ref{tab:1}, we compare the annotation efficiency of three different strategies: (1) AlphaPose*, which uses AlphaPose~\cite{alphapose} to generate initial keypoint predictions followed by manual correction of inaccurate joints; (2) Click-Pose~\cite{yang2023neural}, which predicts initial keypoints and then utilizes its built-in correction mechanism to iteratively refine erroneous points; and (3) IMPose. We recruit 8 participants to annotate 6 video clips, using all three strategies. All annotations are conducted on the same labeling platform with consistent interfaces and workflows. Notably, all participants have prior experience with the platform, having previously contributed to the completion of the PoseTrack21~\cite{doering2022posetrack21}, and are thus familiar with its annotation procedures to varying degrees.

For each strategy, we record the time required to annotate a single person and calculate both the average and standard deviation across all participants. The dataset includes clips from both 3DPW~\cite{von2018recovering} and PoseTrack21~\cite{doering2022posetrack21}, ensuring a diverse range of challenging real-world scenarios. The 6 video clips used in the study all involve varying degrees of occlusion and complex human motions, further reflecting realistic annotation difficulties. Results show that IMPose significantly reduces annotation time in both domains, achieving speedups of approximately \textbf{4} and \textbf{3} times compared to AlphaPose* and Click-Pose, respectively.

\begin{table}[tb]
  \centering
  \imposetablestyle
  \caption{Comparison of the Average and Standard Deviation of Time Cost per Person Annotation via AlphaPose~\cite{alphapose}, Click-Pose~\cite{yang2023neural} and IMPose. Notably, AlphaPose* means that we manually correct poses made by AlphaPose~\cite{alphapose} frame by frame. IMPose achieves the lowest annotation time on both datasets, demonstrating its clear advantage in annotation efficiency.}
  \label{tab:1}
  \ADLinactivate
  \begin{tabularx}{\columnwidth}{l|*{2}{>{\centering\arraybackslash}X}}
    \toprule
    \multirow{2}{*}{Method} & \multicolumn{2}{c}{Time per person (s)$\downarrow$} \\
    \cmidrule(lr){2-3}
    & 3DPW & PoseTrack21 \\
    \midrule
    AlphaPose* & $13.75 \pm 2.82$ & $14.10 \pm 1.74$ \\
    Click-Pose & $10.96 \pm 2.53$ & $10.64 \pm 2.32$ \\
    \rowcolor{imposerow}
    \textbf{IMPose} & $\mathbf{4.13 \pm 0.91}$ & $\mathbf{3.05 \pm 1.13}$ \\
    \bottomrule
  \end{tabularx}
\end{table}

\subsection{Downstream Utility of Corrected Annotations}

To evaluate the training value of IMPose-corrected annotations, we compare original pseudo-labels and corrected annotations for training DSTA and fine-tuning CameraHMR~\cite{patel2025camerahmr}, using the same training samples and configurations within each comparison. DSTA is trained from scratch with the same initialization seed (42), selected using a shared development set, and evaluated on a held-out corrected-label test set containing 130 videos and 102,833 person instances. CameraHMR is fine-tuned with 2D joint reprojection supervision and evaluated against official PoseTrack validation annotations. Both evaluations use ground-truth person boxes and 13 shared keypoints. As shown in Table~\ref{tab:downstream_utility}, corrected annotations reduce DSTA's mean keypoint pixel error (EPE) by 6.27\% and its normalized mean error (NME), normalized by the person-box diagonal, from 5.83\% to 5.38\%. For CameraHMR, corrected supervision improves 2D projection PCKh@0.5 by 3.72 percentage points over pseudo-label supervision and also outperforms the pretrained model. These results demonstrate the value of IMPose-corrected annotations as supervision for downstream keypoint localization.

\begin{table}[t]
  \centering
  \imposetablestyle
  \setlength{\tabcolsep}{2pt}
  \caption{Downstream performance with original pseudo-labels and
  IMPose-corrected annotations. DSTA is trained from scratch, while
  CameraHMR is fine-tuned from the same pretrained model.}
  \label{tab:downstream_utility}
  \ADLinactivate
  \begin{tabularx}{\columnwidth}{l|*{2}{>{\centering\arraybackslash}X}|c}
    \toprule
    \multirow{2}{*}{Setting}
      & \multicolumn{2}{c|}{DSTA}
      & CameraHMR \\
    \cmidrule(lr){2-3}
    \cmidrule(lr){4-4}
      & \shortstack{EPE\\(px)$\downarrow$}
      & \shortstack{NME\\(\%)$\downarrow$}
      & \shortstack{PCKh@0.5\\(\%)$\uparrow$} \\
    \midrule
    Pretrained
      & -- & -- & 78.21 \\
    Pseudo-labels
      & 26.13 & 5.83 & 79.98 \\
    \rowcolor{imposerow}
    \textbf{IMPose-corrected}
      & \textbf{24.49}
      & \textbf{5.38}
      & \textbf{83.70} \\
    \bottomrule
  \end{tabularx}
\end{table}

\section{Future Work and Discussion}
Despite these encouraging results, several directions remain for future research. Although IMPose shows strong robustness in challenging cases, its performance may still degrade under extremely long temporal gaps, prolonged target disappearance, or highly crowded scenes with frequent identity ambiguities. Extending the current 2D framework to 3D human pose and whole-body annotation, as well as integrating stronger foundation models for visual perception and motion correspondence, may further improve the scalability and generalization of interactive annotation systems. We hope IMPose can serve as a practical tool and a strong baseline for future research on efficient and high-quality human pose annotation.
\section{Conclusion}

We present IMPose, a correction-conditioned temporal inference framework for interactive multi-person pose annotation. A correction-aware pose decoder learns to refine imperfect keypoint priors using image evidence. At inference time, propagated human corrections provide geometric priors for target-frame visual refinement. A keypoint-aware identity state preserves person- and joint-level correspondence, while the refined pose and identity state are fed back to subsequent frames to form a closed temporal correction loop.

Experiments on PoseTrack21 and 3DPW demonstrate the effectiveness of IMPose under sparse interaction. On PoseTrack21, three simulated corrections yield 76.15 mAP and 67.25 HOTA, while the manual correction study reports lower annotation time than AlphaPose-based manual correction and Click-Pose. We further extend PoseTrack21 with 187,920 additional person instances through 3,548,968 annotation updates to keypoints, bounding boxes, and person IDs using 100 annotator-hours. Downstream experiments show that the corrected annotations improve DSTA keypoint localization and CameraHMR 2D projection accuracy, supporting their value as training supervision.

\section{ACKNOWLEDGMENTS}
This work was supported in part by the Key Program of National Natural Science Foundation of China (62631020), National Natural Science Foundation of China (62501416), Natural Science Foundation of Tianjin (24JCYBJC01300).

\section*{Statements and Declarations}

\subsection*{Data Availability}
The public datasets used in this study are available from their respective providers. The additional PoseTrack21 annotations generated in this study will be made publicly available.

\subsection*{Competing Interests}
The authors declare no competing interests.

\bibliography{main}

\end{document}